\documentclass[letterpaper, 10 pt, conference]{ieeeconf}  
\usepackage{cite}
\usepackage[ruled,vlined]{algorithm2e}
\usepackage{graphicx}
\usepackage{tcolorbox}
\usepackage{xcolor}
\usepackage{hyperref}
\usepackage[]{footmisc}
\usepackage{adjustbox}
\usepackage{booktabs,graphicx,makecell,siunitx,eqparbox}
\usepackage{tabularx}
\usepackage{footnote}
\usepackage{balance}
\usepackage{subfig}
\usepackage{caption}
\makesavenoteenv{tabular}

\IEEEoverridecommandlockouts                              

\title{\LARGE \bf
Evaluating Uncertainty-based Failure Detection for \\ 
Closed-Loop LLM Planners
}

\author{Zhi Zheng$^{*,1,2}$, Qian Feng$^{1,2}$,  Hang li$^{1,2}$, Alois Knoll$^{2}$, 
Jianxiang Feng$^{*,1,2}$ %
\thanks{*: Equal Contributions, \{zhi.zheng, jianxiang.feng\}@tum.de.}
\thanks{$^{1}$Agile Robot SE}
\thanks{$^{2}$ Department of Informatics, Technical University of Munich}
}

\begin{document}

\maketitle
\thispagestyle{empty}
\pagestyle{empty}

\begin{abstract}


Recently, Large Language Models (LLMs) have witnessed remarkable performance as zero-shot task planners for robotic manipulation tasks. 
However, the open-loop nature of previous works makes LLM-based planning error-prone and fragile. 
On the other hand, failure detection approaches for closed-loop planning are often limited by task-specific heuristics or following an unrealistic assumption that the prediction is trustworthy all the time.
As a general-purpose resoning machine, LLMs or Multimodal Large Language Models (MLLMs) are promising for detecting failures. However, the appropriateness of the "always true" assumption gets exacebated due to the notorious hullucination problem. 
In this work, we attempt to mitigate these issues by introducing a framework for closed-loop LLM-based planning called KnowLoop, backed by an uncertainty-based MLLMs failure detector, which is agnostic to any used MLLMs or LLMs.
Specifically, we evaluate three different ways for quantifying the uncertainty of MLLMs, namely token probability, entropy, and self-explained confidence as primary metrics based on three carefully designed representative prompting strategies. 
With a self-collected dataset including various manipulation tasks and an LLM-based robot system, our experiments demonstrate that token probability and entropy are more reflective compared to self-explained confidence. 
By setting an appropriate threshold to filter out uncertain predictions and seek human help actively, the accuracy of failure detection can be significantly enhanced. 
This improvement boosts the effectiveness of closed-loop planning and the overall success rate of tasks.

\end{abstract}

\section{INTRODUCTION}
Pre-trained Large language models (LLMs) have shown superior generalization capability as zero-shot task planners in robot learning by transforming high-level language instructions into low-level action plans~\cite{pmlr-v162-huang22a, ahn2022i,driess2023palme,huang2023voxposer,liang2023code}. 
However, these planners predominantly employ open-loop control, rigidly following initial plans without incorporating environmental feedback. 
Consequently, execution errors or plan deficiencies remain unaddressed, potentially leading to task failure.

Recent works on facilitating closed-loop planning for LLM planners tend to be less attentive toward the indispensable first step in closed-loop planning -- failure detection~\cite{raman2022planning, skreta2023errors, liu2023reflect, guo2023doremi}.
They usually adopt either LLMs~\cite{raman2022planning, skreta2023errors, liu2023reflect} or Multimodal Large Language Models (MLLMs)~\cite{guo2023doremi} by assuming their predictions to be trustworthy all the time.

\begin{figure}[t]
\includegraphics*[width=\linewidth]{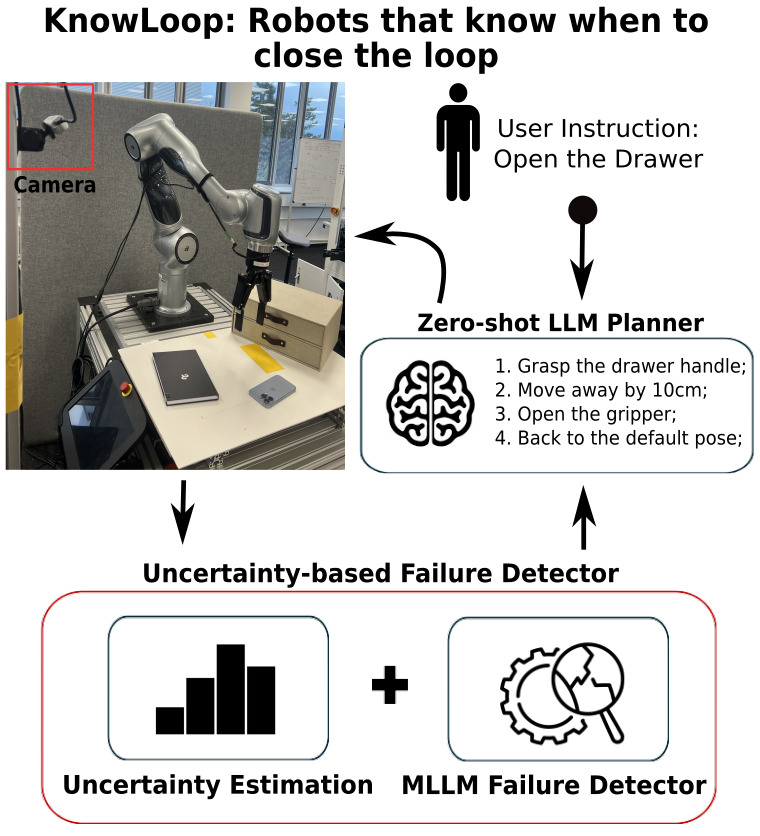}
\centering
\caption{A Sketch of our proposed closed-loop planning framework, KnowLoop featured by an uncertainty-based failure detection module, allowing robots to know when to close the loop properly during planning.}
\label{fig:teaser}
\vspace{-25pt}
\end{figure}

This simplified assumption might not hold due to the notorious hallucination problem of LLMs/MLLMs~\cite{huang2023survey}, meaning the inconsistencies between the generated contents and the factual truth or user input contexts. 
On the other hand, some heuristic-based failure detection~\cite{huang2022inner} for LLM planners is relatively accurate but highly task-specific, restricting their general suitability to various tasks and models. 
Uncertainty estimation~\cite{gawlikowski2023survey,pmlr-v119-lee20b} has been considered a promising technique to handle the hallucination issue for LLMs~\cite{huang2023survey} and found widespread adoption in robotics~\cite{firoozi2023foundation, pmlr-v164-lee22c,feng2022bayesian,feng2019}.
More noteworthy, uncertainty estimation for LLMs benefits from its characteristic of being model-agnostic, hence facilitating wider suitability.

In light of this, we introduce KnowLoop (see Fig. \ref{fig:teaser}), a framework backed up by an uncertainty-based failure detection module, allowing robots to know when to close the loop properly during planning.
The proposed module is LLM/MLLM model agnostic, meaning more general applicability when compared with the task-specific approaches. 
Meanwhile, it is able to boost the performance of LLM-based closed-loop planning by filtering out uncertain predictions from the failure detector.
When confronting predictions with low confidence, the framework enables the robot to solicit human help actively in order to avoid task failure during task execution.
Specifically, we first estimate the uncertainty for a MLLM, which we adopt as a failure detector given the task description and an image depicting the current stage.
Inspired by~\cite{ren2023selfevaluation}, we evaluate three approaches for obtaining uncertainty estimates in MLLMs, namely token probability, entropy, and self-explained confidence.
In addition, to comprehensively investigate the impact of different prompts to trigger the MLLM-based failure detector, we carefully design three prompting strategies spanning over two representative categories.
The first category is direct prompts, in which the question of success or failure appears, while the second one is indirect prompts, where there is no this kind of question in the context.  

\begin{figure*}
    \centering
    \includegraphics[scale=0.15]{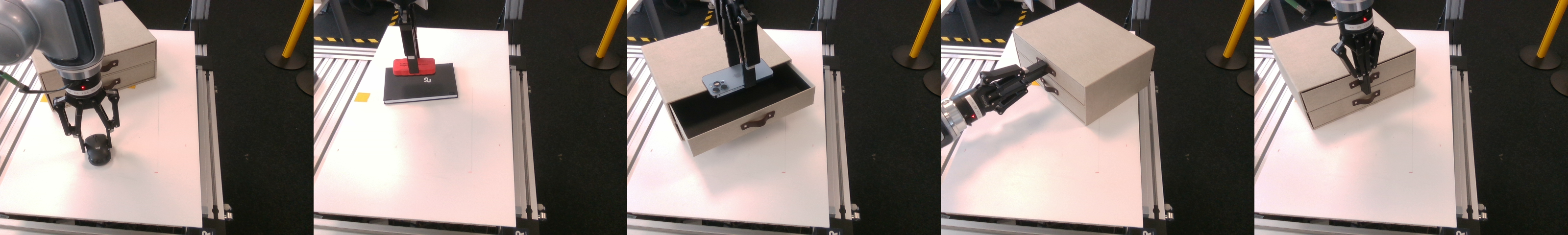}
    \caption{Example Images for different tasks in the self-collected dataset, from left to right: picking up a mouse, placing a sponge on the notebook, placing a smartphone in the drawer, opening the drawer, and closing the drawer.}
    \label{fig: dataset_img}
    \vspace{-10pt}
\end{figure*}

To study the effectiveness of the proposed idea, we build an LLM-based robot system adapted from ~\cite{huang2023voxposer} and create a self-collected dataset for success/failure analysis with five different manipulation tasks, including both short-term and long-term complex actions, see Fig. \ref{fig: dataset_img}.
We verify the benefits of exploiting uncertainty-based failure detection for closed-loop LLM planning through empirical experiments. 
To assure comprehensiveness, the experiments are conducted on both the dataset and real hardware with two different MLLMs (LLaVA~\cite{liu2023visual} and ChatGPT-4V).
The contributions made in this work are as follows:
\begin{itemize}
    \item We propose to exploit uncertainty estimation for more trustworthy and effective failure detection based on LLMs/MLLMs.
    \item We integrate the uncertainty-based MLLM failure detector into a zero-shot LLM-based robot system to facilitate closed-loop planning. 
    \item With experiments on a self-collected dataset and real hardware, we show promising results of leveraging uncertainty estimates in LLM/MLLM for closed-loop planning.
\end{itemize}

\section{RELATED WORK}
To lay the groundwork for a comprehensive understanding, we review the literature in the following relevant areas:

\subsubsection{Uncertainty Estimation in Robotics}
Uncertainty estimation has a longstanding tradition in robotics, dating back to the era of probabilistic probiotics~\cite{thrun2002probabilistic} and robotic introspection~\cite{grimmett2016introspective}.
With the advent of deep learning, uncertainty estimation is gaining more and more attention in achieving trustworthy learning-enabled robotics due to the black-box nature of neural networks~\cite{sunderhauf2018limits,gawlikowski2023survey,feng2023topologymatching}.
The spectrum of applying uncertainty estimation to empower robots spans from learning-based perception~\cite{feng2019,feng2022bayesian,pmlr-v164-lee22c} to control~\cite{loquercio2020general}.
Moreover, the significance of uncertainty estimation is increasingly prominent in the age of foundational models~\cite{firoozi2023foundation}.
When it comes to LLMs for robotics, there is only a limited amount of work, such as KnowNo~\cite{ren2023robots}, which proposes a method to estimate the uncertainty of an LLM planer.
It is done by generating multiple options as the next steps and analyzing their corresponding token probabilities with conformal prediction. 
However, they do not consider the failure of task execution itself. 
\subsubsection{Uncertainty Estimation in LLMs}
LLMs have shown an inevitable tendency to create hallucinations, which refers to the inconsistencies between the generated contents and the real-world truth or user inputs~\cite{huang2023survey}.
The model's uncertainty is demonstrated to be related to the occurrence of LLM hallucinations~\cite{varshney2023stitch}. 
Using uncertainty estimation to predict LLM hallucination as a zero-resource setting prevents the need for external knowledge resources.
\cite{xiong2024llms} attempt to use Prompt Engineering to have the LLM output the confidence level of the answer simultaneously when speaking it out, as well as obtaining the model's confidence level based on the Consistency base method.
~\cite{ren2023selfevaluation} propose an LLM self-evaluation method in the form of multiple-choice questions or true/false statements to obtain a quality-calibrated confidence score at the token level. 
~\cite{xiong2024llms} introduce a confidence elicitation framework consisting of prompting, sampling, and aggregation strategies. The framework is evaluated with confidence calibration and failure prediction tasks. Their focuses are both on the general question-answer settings instead of robotic applications.


\subsubsection{Closed-loop planning with LLMs}
LLMs can be harnessed to translate high-level, abstract task instructions into actionable, step-by-step sequences for execution by agents. Recent studies~\cite{huang2022inner,yao2023react,wang2023describe,shinn2023reflexion,sun2023adaplanner} have showcased the capabilities of LLMs for self-reflection and correction in response to environmental feedback. Here they often assume a ground truth environmental feedback. 
A prior work similar to ours~\cite {liu2023reflect} utilizes multisensory data and hierarchical summary to detect, reason, and correct failures in closed-loop robotic task planning in real-world scenarios. 
They rely on dedicated hand-crafted external failure detectors but we propose to estimate the uncertainty from a multimodal Large Language Model (MLLM) for more general failure detection, providing a more versatile and adaptable solution.

\begin{figure*}[htbp]
    \centering
    \includegraphics[width=\textwidth, height=8cm]{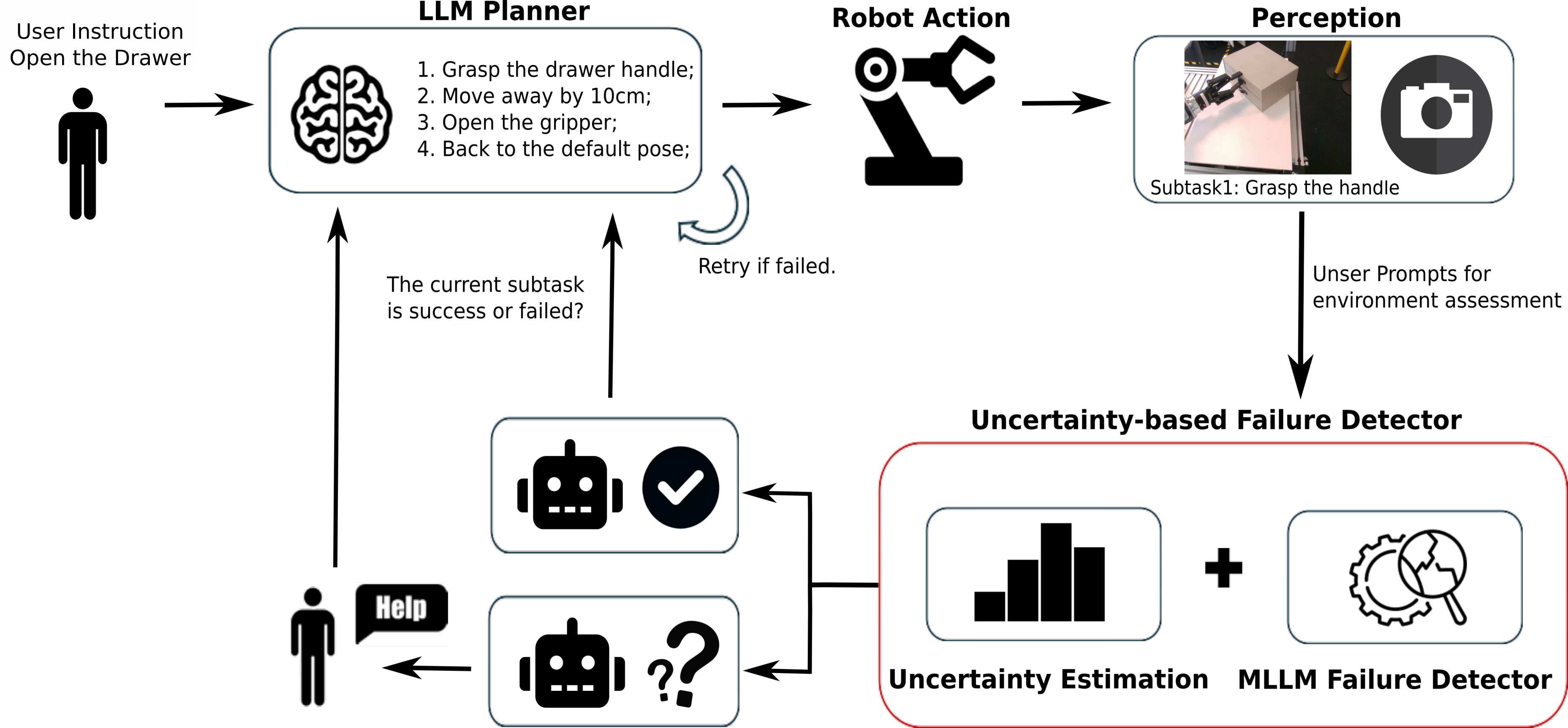}
    \caption{A flow diagram of the proposed framework for the task: opening the drawer.}
    \label{fig:enter-label}
    \vspace{-15pt}
\end{figure*}
\section{The Proposed Closed-Loop Planner}
\label{sec:Method}

In this chapter, we outline KnowLoop, the proposed closed-loop control system's framework. 
Our implementation of an interactive failure detector is based on uncertainty derived from the MLLM. 
Specifically, we utilize open-source MLLMs such as LLaVA \cite{liu2023visual} and ChatGPT-4V for failure detection within an LLM planner based on ChatGPT-4. 
Note that the LLMs/MLLMs used in this work can be merged into one in principle, which we leave for future work. 
The procedure entails a decomposition of a language instruction into a series of sub-goals. 
After executing each sub-goal, the failure detector is activated to collect feedback from environment images and determine whether the sub-goal has been \textit{achieved}.
In the following, we will articulate the major components of the proposed uncertainty-based failure detector, including the prompting strategies to activate the MLLM failure detector and approaches for uncertainty quantification in LLMs.
Finally, we integrate the uncertainty-based failure detector into an LLM-based planning framework for closed-loop planning.

\subsection{Uncertainty-based Failure Detection}
A proper and informative prompt is necessary to trigger an MLLM failure detector.
For this purpose, we specifically design two representative categories of prompting strategies, namely a direct one and an indirect one. 
Direct prompting strategy directly entails the question of success or failure, while indirect one does not encompass this kind of direct question of failure or success in the context.  
Furthermore, to tackle the improper assumption of the failure detector, we consequently estimate the uncertainty of each response to these prompts. 
Based on the uncertainty estimates and a predetermined threshold, the \textit{reliability} of each response can be evaluated quantitatively in a generic way.
Responses with excessively high uncertainty will be filtered out, and instead, human feedback will be requested to assess the current state, as shown in Algorithm \ref{alg:failure_detect}.
More noteworthy, the proposed way to detect failure cases is agnostic to LLM used in the planner.
Therefore, it enjoys the merit of general applicability compared to other task-specific or heuristic approaches. 

\subsubsection{Prompting Strategies}
The impact of utilizing diverse prompts for an LLM is substantial, as the chosen prompt directly shapes the model's output in manifold ways. 
To attain improved detection outcomes and reasonable uncertainty, it is crucial to meticulously select the prompts employed in the detection process. 
Toward this objective, we have carefully designed three distinct prompt strategies, articulated as follows.

\textbf{Direct Prompts via Subgoal State Comparison (SSC).}
Inferring the state after an action and comparing it to the environmental images is challenging for LLMs. To address this, we propose outlining each subgoal or action, along with its expected state description. The MLLM will then compare this with the current environmental image to determine if the present state aligns with the anticipated outcome, indicating the successful achievement of the subgoal.

\textbf{Direct Prompts via Spatial Relationship Analysis (SRA).}
The "Chain of Thought" (CoT)~\cite{cot} reasoning in LLMs epitomizes a sophisticated strategy for tackling complex problem-solving tasks. 
The technique utilizes the model's ability to generate intermediate steps or reasons, mimicking human reasoning. In robot manipulation tasks, notable changes occur in the interaction between the gripper and the spatial arrangement of objects after each action. Rather than solely assessing success or failure based on state descriptions and images, the MLLM should involve an intermediate cognitive process that compares current and expected states to determine the action's outcome.

\textbf{Indirect Prompts via Next Action Prediction(NAC).}
To differentiate this approach from previous ones, success is no longer a simple yes or no. Instead, after each action, we present all high-level plans and corresponding executable actions to the MLLM. We then provide images to illustrate the current environment. Based on this, the MLLM selects the next action from the options presented, turning the detection problem into a multiple-choice scenario. Success occurs when the MLLM's selection aligns with the next step of the predefined plan, while any alternative action is considered a failure.

\subsubsection{Uncertainty Quantification}
The failure detector's accuracy is crucial for the closed-loop control system's success. True negatives can cause the system to miss errors, leading to task failure, while false negatives can disrupt operations. Even advanced MLLMs like ChatGPT-4V may produce \textit{hallucinating} answers, and it's important to gauge their confidence. When uncertain, it's better to say "I don't know" than provide a false response.

\begin{tcolorbox}[title=Subgoal State Comparison (SSC)]
The robot arm is given a task: \textcolor{red}{[task instruction]}. The robot arm just tried to execute \textcolor{red}{[subtask at time t]}.
\\
Q: Based on the image, is the \textcolor{red}{[expected state description]} satisfied? 
\\
A: Yes / No.
\end{tcolorbox}
\begin{tcolorbox}[title=Spatial Relationship Analysis (SRA)]
The robot arm is given a task: \textcolor{red}{[task instruction]}. The robot arm just tried to execute \textcolor{red}{[subtask at time t]}.
\\
Q: To tell whether \textcolor{red}{[expected state description]} is satisfied, first analyze the spatial relationship between objects in the working space. 
\\
A: [analysis]
\\
Q: Is the \textcolor{red}{[expected state description]} satisfied? 
\\
A: Yes / No.
\end{tcolorbox}

\begin{tcolorbox}[title=Next Action Prediction (NAC)]
The robot arm is given a task: \textcolor{red}{[expected state description]}. The high-level plan for this task is \textcolor{red}{[list of subtasks]}. The robot arm just tried to execute \textcolor{red}{[subtask at time t]}.
\\
Q: Based on the image, which subtask should be the next step? 
\\
A: B (One of the subtasks in the list).
\end{tcolorbox}

\textbf{Token probability.}
Prior work~\cite{ren2023robots} suggests that, by formulating multiple-choice questions and answering (MCQA) and having LLMs' responses to them, it is feasible to deduce the occurrence probability of each option token $ P(x_i)$. 
This probability is interpreted as the likelihood of the corresponding option being correct. 
Deriving the token probability from the MLLM output can act as an indicator of the probability of affirmative or negative responses.

\textbf{Entropy.}
In information theory, entropy represents a measure of uncertainty or disorder. 
Within the output layer of the MLLM, each token is assigned a score. 
The probability of a token can thus be interpreted as the probability of a given classification. 
The corresponding entropy may reflect the model's uncertainty about its response to a certain extent.

\textbf{Self-explained Confidence.}
In certain instances, LLMs can be induced to articulate their confidence levels explicitly. 
For example, the model might be prompted to deliver responses in the format: "I am X\% certain that the answer is Y," wherein X represents the model's \textit{self-evaluated confidence level}. 
This approach is contingent upon meticulous prompt engineering. 
Given that it hinges on the model's inherent processes for comprehending and generating confidence-related responses, it may not consistently produce precise confidence estimations.

\subsection{Closed-loop Planning}
In the final proposed closed-loop control framework, ChatGPT-4 is used as the high-level task planner.
The LLM planner understands language instructions and decomposes them into multiple subtasks. 
GPT's code generation function will be called for each subtask, and combined with the robotic arm motion API, it will generate corresponding Python code for each action. 
Subsequently, the robotic arm will execute actions one by one according to the resultant task plan. 
After each action execution, the detector will be called to assess the state of the environment and detect the failure. 
If the assessment indicates successful execution, the robotic arm will proceed with the following action as planned. 

\begin{algorithm}
\SetAlgoLined
\KwIn{Prompts of the current subtask $i$: $l_i$, Image of the current state of step $i$: $R_i$, a user-specified threshold: $\delta$.}
\KwOut{Whether the subtask was successfully excuted}
\SetKwFunction{FMain}{failure\_detect}
\SetKwProg{Fn}{Function}{:}{}
\Fn{\FMain{$l_i$, $R_i$}}{
    uncertainty\_est, response = MLLM($l_i$, $R_i$)\;
    \uIf{uncertainty\_est $< \delta$}{ 
        \KwRet{response}\;
    }
    \Else{
        \textit{user\_help} $\leftarrow$ \texttt{input("I am not sure! The current subtask is successful or failed? ")}\;
        \KwRet{user\_help}\;
    }
}
 \caption{Failure Detection.}
\label{alg:failure_detect}
\end{algorithm}
If the assessment indicates failure, the robotic arm will stop the current action and re-execute all actions as shown in Algorithm \ref{alg:close-loop}.
\begin{algorithm}
\SetAlgoLined
\KwIn{A task planning prompt: $L$,  maximum number of retrying: $k$.}
\SetKw{KwLLM}{LLM Code Generation based on $L$:}

\KwLLM

$subtask\_list = [l_1, l_2, \ldots, l_n]$\;
  $retry = 0$\;
  $index = 0$\;
  \While{index $<$ len(Subtasks) and retry $<$ $k$}{
   $l_i = subtask\_list[index]$\;
   $R_i = robot\_execution(l_i)$\;
   $prediction = failure\_detect(l_i, R_i)$\;
   \eIf{prediction is success}{
    $index = index + 1$\;
    }{
    $index = 0$\;
    $retry = retry + 1$\;
   }
  }
 \caption{Closed-loop Planning Framework.}
\label{alg:close-loop}
\end{algorithm}

\section{Experiment}
We will first present our experimental setup and implementation details, followed by a discussion of the experiment results. Then, we will evaluate three uncertainty estimation methods and select the best-performing combination for closed-loop LLM planners on real hardware based on the results.
To note that the test set includes slightly different tasks\footnote{Task\_1:pick up the sponge and place it on the notebook, Task\_2:pick up the smartphone and place it in the upper drawer, Task\_3:open the upper drawer} but with similar items. 
In interference-free scenarios, the robotic arm successfully completes tasks $70-80\%$ of the time. When human interference is introduced—such as displacing objects from their intended locations or impeding operations, the success rate drops to $0\%$ under open-loop control conditions.
\subsection{Dataset}
Valid uncertainty implies that higher model uncertainty correlates with decreased prediction accuracy. 
Conversely, as the model's confidence increases, its accuracy significantly improves. 
Additionally, the applicability of this uncertainty in robotic manipulation tasks remains uncertain. 
To investigate this more efficiently, we collected 142 samples representing different execution states across five tasks\footnote{Task\_1:pick up the mouse and place it on the notebook, Task\_2:pick up the sponge and place it in the upper drawer, Task\_3:pick up the smartphone and place it on the drawer, Task\_4:open the upper drawer, Task\_5:close the upper drawer} in Table \ref{tab:dataset}. 
The primary failures encompass detection and operational errors, explicitly excluding planning errors. The tasks primarily involve actions such as grasping, placing, and pushing. 

\subsection{Evaluation Metrics}
In this subsection, we explain the metrics used for evaluating different variants in the experiment:
\textbf{a) Uncertainty-Acuracy Curve and 
Area Under Curve (AUC) (Calibration-AUC)}~\cite{ren2023selfevaluation} depicts accuracy as a function of the abstention uncertainty threshold. 
This threshold determines the point at which samples are excluded from consideration if their uncertainty surpasses $(1-threshold)$. 
Initially, no samples are omitted when the uncertainty threshold is set to zero, allowing the accuracy measurement to reflect the entire dataset's performance. 
As the uncertainty threshold escalates, samples exhibiting uncertainty above $(1-threshold)$ are selectively discarded. 
Subsequently, the accuracy metric is recalculated to reflect the composition of the residual dataset. 
Should uncertainty serve as a reliable indicator of the model's uncertainty, discarding samples marked by higher uncertainty should theoretically enhance the quality of the retained samples. 
Consequently, this selective exclusion is expected to improve the model's overall accuracy. 
\textbf{b) Selective Generation Curve and AUC (Selective-AUC)}~\cite{ren2023selfevaluation} depits accuracy as a function of the abstention rate, denoted as $\alpha$. 
This approach involves ordering the samples by their uncertainty and then abstaining from the top $\alpha\%$ of samples based on their uncertainty values. 
No samples are abstained at $\alpha = 0 \% $, meaning the curve initiates at the conventional accuracy metric. 
Consequently, an increase in the curve is anticipated as $\alpha$ escalates. 
\textbf{c) Success Rate}
is calculated by dividing the number of successful experiments by the total number of experiments.
\textbf{d) Detection Accuracy} is the number of correct predictions divided by the total number of detections made by the MLLM failure detector. 
Instances where humans provide assistance are not included. 
This metric indicates the ability of MLLM to make accurate predictions.
\textbf{e) Human Involve Rate} is calculated by dividing the number of times humans provided assistance by the total number of detections performed during the experiment. 
A higher human involvement rate indicates a larger effort to achieve a high success rate, which further implies the limited performance of the MLLM failure detector.
\textbf{f) Generation Rate} only applies to Self-explained Confidence, as not every attempt results in a successful generation of confidence scores. 
It is calculated as the number of successful confidence score generations divided by the total number of tests.
\subsection{System Setup}
We use a Diana 7 robot arm and Robotiq 2f-140 gripper with a tabletop setup (see Fig. \ref{fig:teaser}). 
We mount one eye-to-hand Realsense D415 RGBD camera on the top left from the top-down view.
The camera returns the real-time RGB-D observations at 30 Hz.
Our LLM-based planner is adapted based on Voxposer~\cite{huang2023voxposer}. 
We use Grounding-Dino~\cite{GroundingDino} for open-set object detection, then feed it into Segment Anything~\cite{SAM} to obtain a mask. 
The mask is used to crop the object point cloud.
We run our system on a single Nvidia RTX-4090 GPU.
\begin{table}[h]
\vspace{-15pt}
\caption{Self-Collected Dataset Break Down}
\begin{center}
\label{tab:dataset}
\begin{tabular}{|c|c|c|c|c|c|c|}
\hline
 & total &task\_1 & task\_2 & task\_3 & task\_4 & task\_5\\
\hline
success & 72 & 12 & 18 & 12 & 18 & 12  \\
\hline
failure & 70 & 14 & 19 & 10 & 14 & 13  \\
\hline
\end{tabular}
\end{center}
\vspace{-20pt}
\end{table}

\subsection{Results}
\begin{table*}[ht!]
\centering
\caption{Results Comparison of three uncertainty estimation methods and three prompting strategies on the self-collected dataset based on LLaVA. $^1$: Calibration-AUC is not computable for self-explained confidence as the MLLM might fail to provide a sensible response. Instead, we provide the generation rate.}
\begin{center}
\label{tab:uncertainty curve} 
\begin{adjustbox}{width=0.95\textwidth}
\begin{tabular}{r|rrr|rrr|rrr}
& \multicolumn{3}{c}{Token Proabbility} & \multicolumn{3}{c}{Entropy} &\multicolumn{3}{c}{Self-explained Confidence$^1$} \\
\midrule
\makecell{Prompting \\Strategies} &\makecell{Detection  \\ Accuracy} & \makecell{Calibration-\\AUC} & \makecell{Selective-\\AUC} &  \makecell{Detection  \\ Accuracy} & \makecell{Calibration-\\AUC} & \makecell{Selective-\\AUC} &  \makecell{Generation\\Rate} & \makecell{Detection  \\ Accuracy}  &\makecell{Selective-\\AUC} \\ 
\midrule
SSC & 57.5\% & 0.58 & 0.65 & 57.5\% & 0.67 & 0.65 & 46.2\% & 57.1\% &0.44 \\
SRA & 66.0\% & 0.64 & 0.68 & 66.0\% & 0.66 & 0.68 & 68.9\% & 59.0\% &0.50 \\
NAP & 29.2\% & 0.58 & 0.49 & 29.2\% & 0.49 & 0.44 & 58.5\% & 3.2\% &0.04\\
\end{tabular}
\end{adjustbox}
\vspace{-20pt}
\end{center}
\end{table*}

\begin{table*}[ht!]
\centering
\caption{Results of uncertainty-based failure detection for closed-loop LLM Planners on real hardware. Entropy is used as the uncertainty measure for both ChatGPT-4V and LLaVA.}
\begin{center}
\label{tab:close_loop} 
\begin{adjustbox}{width=0.8\textwidth}
\begin{tabular}{r|rr|rr|rrr}
& \multicolumn{2}{c}{ChatGPT-4V} & \multicolumn{2}{c}{LLaVA} &\multicolumn{3}{c}{KnowLoop (LLaVA+Human Help)} \\
\midrule
Tasks &\makecell{Success\\Rate} & \makecell{Detection  \\ Accuracy}&  \makecell{Success \\Rate} & \makecell{Detection  \\ Accuracy} &  \makecell{Success \\Rate} & \makecell{Detection  \\ Accuracy} & \makecell{Human Involve \\ Rate} \\ 
\midrule
Task\_1 & $10\%$ & $38\%$ & $30\%$ & $57.6\%$ & $70\%$ & $78\%$ & $28\%$ \\
Task\_2 & $20\%$ & $35.4\%$ & $50\%$ & $57.8\%$ & $80\%$ & $81.4\%$ & $31.8\%$ \\
Task\_3 & $40\%$ & $34.1\%$ & $10\%$ & $25.8\%$ & $70\%$ & $76\%$ & $21.8\%$\\
\end{tabular}
\end{adjustbox}
\vspace{-20pt}
\end{center}
\end{table*}

\subsubsection{Uncertainty Estimation for MLLM Failure Detecor}
\begin{figure}[!ht]
    \centering
    %
    \subfloat[]
    {\includegraphics[width=0.5\linewidth]{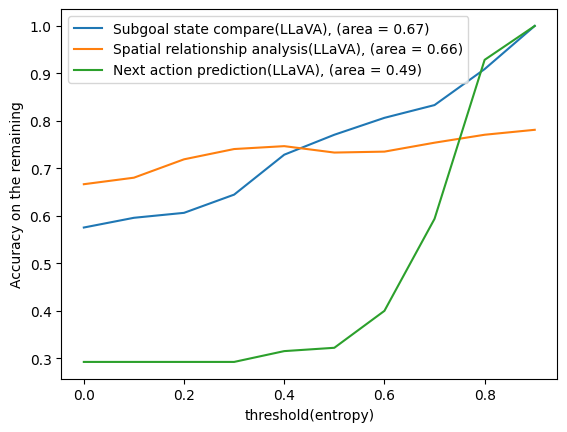}}
    \subfloat[]
    {\includegraphics[width=0.5\linewidth]{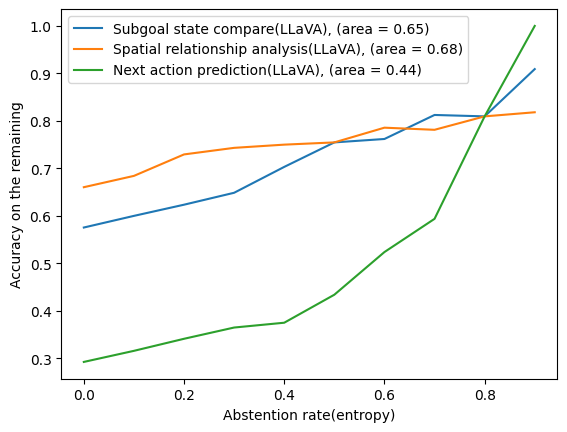}}
    \\
    \subfloat[]
    {\includegraphics[width=0.5\linewidth]{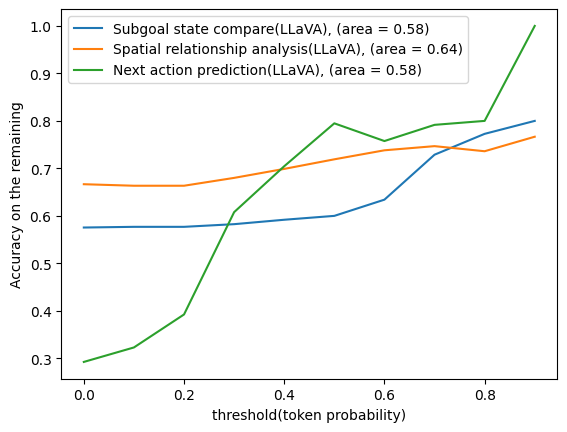}}
    \subfloat[]
    {\includegraphics[width=0.5\linewidth]{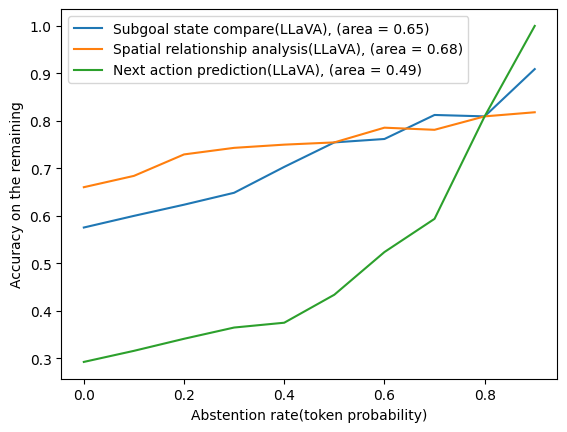}}
    \caption{Uncertainty Accuracy Curve (left column) and Selective Generation Curve (right column) for three prompting strategies based on entropy (top row) and token probability (bottom row).}
    \label{fig:token_entropy}
    \vspace{-20pt}
\end{figure}

We first evaluate the MLLM failure detector based on LLaVA, as we need token probability for uncertainty calculation which is not supported by ChatGPT-4V API at the moment.
We employ the following criteria to assess the correlation between the uncertainty derived from various prompting and quantification methods and the accuracy of the predictions shown in Table \ref{tab:uncertainty curve}, Fig. \ref{fig:token_entropy}.

The increasing curves in Fig. \ref{fig:token_entropy} indicate that uncertainty quantification through token probability and entropy yields promising performance. 
A distinct trend emerges, showing increased accuracy as uncertainty diminishes. 
However, the correlation between self-explained uncertainty and accuracy is weak, rendering this measure impractical for use. 
In Table \ref{tab:uncertainty curve}, the AUC values for the strategies of SSC and SRA are relatively similar and notably superior to that of the NAP strategy. 
This disparity is primarily attributed to the lower accuracy of the NAP strategy in the absence of filtering. 
The linear correlation between accuracy and entropy-based uncertainty is most noticeable in the SSC prompt. We have chosen to use the curve generated by the SSC prompting strategy with entropy for the real hardware experiment due to the promising performance of SSC with entropy.




\subsubsection{Uncertainty for Closed-Loop LLM Planners}
In the real-world experiment, we chose ChatGPT-4V and LLaVA as baselines to compare with the proposed framework KnowLoop. 
In both experiments with these MLLMs, the detector fully trusts the MLLM's prediction results without considering uncertainty. 
For KnowLoop, we filter the failure predictions based on uncertainty with a threshold of 0.6, above which the corresponding prediction will \textit{not} be trusted, and the model will actively seek human help for failure detection. 
In Table \ref{tab:close_loop}, when compared to the baselines that completely trust the MLLM's predictions, our framework can, to some extent, improve the task success rate by around 50\% compared to ChatGPT-4v and by around 40\% compared to LLaVA.
The promising results demonstrate that KnowLoop can enable the robot to actively ask for human help when the prediction is uncertain. 

\section{CONCLUSION}
In this study, we have demonstrated that uncertainty quantified by token probability, and entropy is able to reflect the predictive quality of the MLLM LLaVA. 
We further integrate this into an MLLM failure detector within a zero-shot LLM planner to facilitate closed-loop planning -- a framework dubbed KnowLoop. 
With this, lower-quality responses can be effectively identified using a threshold. 
Our proposed framework can mitigate model illusions or overconfidence, thereby enhancing model accuracy. 
Such improvements render it suitable for the detection phase of closed-loop control systems.
For future work, it would be more resource-efficient to use the MLLM to substitute the LLM in the task planner.
Moreover, conformal prediction can be employed to determine the threshold in the framework with a statistical guarantee~\cite{ren2023robots}.
In addition, investigating how uncertainty estimation can be utilized for failure reasoning and correction is an important next step in setting up a robust and effective closed-loop system.










\balance
\bibliographystyle{IEEEtran}
\bibliography{chapter/reference}

\end{document}